\documentclass[final]{cvpr}
\usepackage{bm}
\usepackage{soul}
\usepackage{cuted}
\usepackage{times}
\usepackage{bbold}
\usepackage{epsfig}
\usepackage{xcolor}
\usepackage{amsmath}
\usepackage{amssymb}
\usepackage{comment}
\usepackage{caption}
\usepackage{colortbl}
\usepackage{arydshln}
\usepackage{graphicx}
\usepackage{booktabs} 
\usepackage{algorithm}
\usepackage{tabu}
\usepackage[accsupp]{axessibility}
\usepackage[numbers,sort]{natbib}
\usepackage{setspace}

\DeclareCaptionFormat{myformat}{#3}

\newcommand{\app}[0]{{Appendix}}

\usepackage{pifont}
\newcommand{\cmark}{\ding{51}}%
\newcommand{\xmark}{\ding{55}}%

\usepackage{xpunctuate}

\DeclareMathOperator*{\argmax}{\arg\!\max}

\definecolor{mygreen}{RGB}{34,139,34}
\definecolor{myblue}{rgb}{0.19, 0.55, 0.91}
\definecolor{googleblue}{RGB}{66,133,244}
\definecolor{googleorange}{RGB}{255,171,64}

\usepackage[pagebackref=true,breaklinks=true,colorlinks,bookmarks=false]{hyperref}
\usepackage{cleveref}

\def\cvprPaperID{2107} 
\def\confYear{CVPR 2022}

\begin{document}

\title{Temporal Alignment Networks for Long-term Video}


\author{%
  Tengda Han$^1$ \quad Weidi Xie$^{1,2}$ \quad Andrew Zisserman$^1$ \\
  $^1$Visual Geometry Group,  University of Oxford \hspace{5pt}
$^2$Shanghai Jiao Tong University\\
  \normalsize{\href{https://www.robots.ox.ac.uk/~vgg/research/tan/}
  {\texttt{https://www.robots.ox.ac.uk/{\textasciitilde}vgg/research/tan/}}} \\  
}

\maketitle

\begin{abstract}
\vspace{-4mm}
The objective of this paper is a temporal alignment network 
that ingests long term video sequences, and associated text sentences, 
in order to: 
(1) determine if a sentence is alignable with the video; 
and (2) if it is alignable, then determine its alignment.
The challenge is to train such networks from large-scale datasets, 
such as HowTo100M, where the associated text sentences have significant noise, 
and are only weakly aligned when relevant.

Apart from proposing the alignment network, 
we also make four contributions:
(i) we describe a novel co-training method 
that enables to denoise and train on raw instructional videos 
without using manual annotation, 
despite the considerable noise;
(ii) to benchmark the alignment performance, 
we manually curate a 10-hour subset of HowTo100M, totalling 80 videos, 
with sparse temporal descriptions.
Our proposed model, trained on HowTo100M, 
outperforms strong baselines~(CLIP, MIL-NCE) on this alignment dataset by a significant margin;
(iii) we apply the trained model in the zero-shot settings to multiple downstream video understanding tasks 
and achieve state-of-the-art results, including text-video retrieval on YouCook2, 
and weakly supervised video action segmentation on Breakfast-Action;
(iv) we use the automatically-aligned HowTo100M annotations for
end-to-end finetuning of the backbone model, and obtain
improved performance
on downstream action recognition tasks.
\end{abstract}

\vspace{-6mm}
\section{Introduction}\label{sec:intro}
\vspace{-1mm}
The recent CLIP and ALIGN papers~\cite{Radford21,Jia21} have demonstrated that 
a combination of large scale paired image-caption data, 
and a simple noise contrastive learning loss can be used to learn powerful image-text embeddings from scratch. 
The image-caption data can be crawled from the internet at scale, for example from image alt-text, 
and the resulting embeddings demonstrate strong ``zero-shot'' generalization abilities.
In the video domain, 
there also exists large-scale sources of text supervision, 
\emph{e.g.}~narrated instructional videos such as the  HowTo100M~\cite{Miech19} dataset,
where demonstrators explain their actions while performing a complex task.
The narrations are unconstrained and can be combinatorially complex, 
including information on ``what'', ``where'' and ``when'',  
such as the actions, the objects, human-object interactions, {\em etc}.

\begin{figure*}[t]
    \centering
	\includegraphics[width=\textwidth]{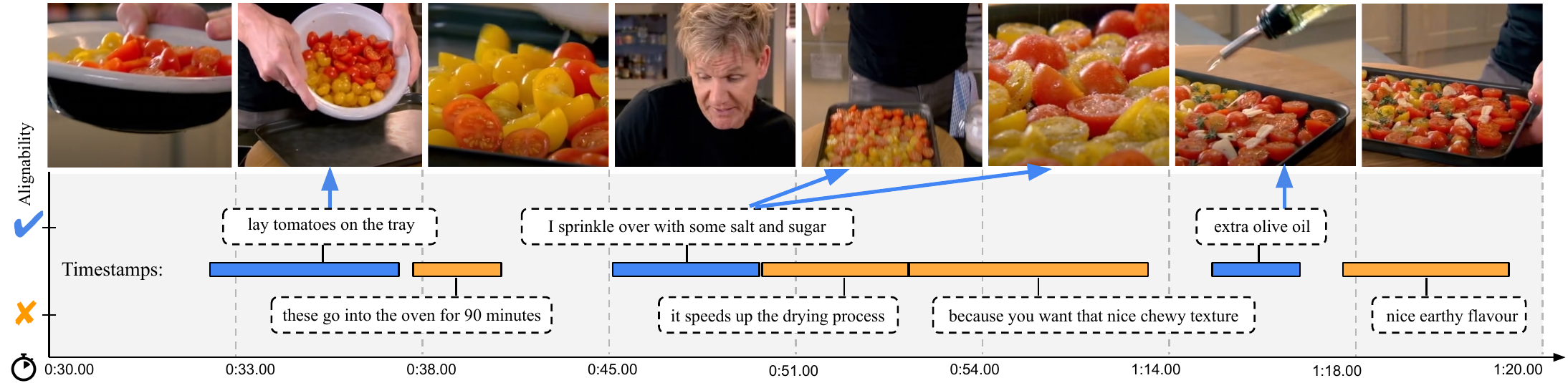}
	\vspace*{-7mm}
	\caption{
	\footnotesize{An example of visual-textual mis-alignment in a raw instructional video.
	The presenter's narration can be not visually relevant at all, 
	{\em e.g.}~describing a flavor;
	or asynchronous with visual content by a time difference.
	The \textcolor{googleblue}{\textbf{\cmark}} and \textcolor{googleorange}{\textbf{\xmark}} 
	indicate visually alignable and non-alignable text, respectively (by human judgement). 
	The colored bar shows the start-end timestamp of narration. 
	Example from \scriptsize{\url{https://www.youtube.com/watch?v=M8OGXmLTTiI?t=30}}.}}
\vspace{-14pt}
\label{fig:align}
\end{figure*}

However, these instructional videos pose additional fundamental challenges 
over the image-caption scenario due to the temporal \emph{alignment} problem~(illustrated in Figure~\ref{fig:align}): 
(i) the demonstrator often makes statements that are unrelated to the visual signal, 
such as describing food taste or explaining the consequence of actions.
These texts are {\em not visually alignable}.
(ii) the demonstrator might explain their action before or after performing it,
and their statements often {\em do not follow the same order} as their actions,
resulting in the text and visual entities being asynchronous.
These texts are  {\em not temporally aligned} to the visual signal.
Additionally, unlike spatial segmentation in images, 
where objects boundaries are often formed by a discontinuity between regions with strong gradients, 
temporal actions in videos are often continuous, 
making it difficult to clearly define the start and end points for the temporal interval.
Last but not the least, 
there is additional noise coming from the imperfect 
Automatic Speech Recognition (ASR) systems on the spoken narrations.
Note that the image-caption data does not face these problems since captions are provided by human annotators for that image; 
although they may be incomplete, there is no temporal alignment issue.

The extent of these alignment challenges is significant~\cite{Miech19,Miech20}.
In 10 hours of instructional videos~(sourced from HowTo100M) 
that we annotated for this work, 
only 30\% of the narration sentences are visually alignable, 
and only {15\%} are naturally well-aligned.
This means that 
the demonstrator is describing their action 
synchronously with the video only 15\% of the time.
If the alignment issues are resolved then the benefits of learning 
from such narrated instruction videos can potentially be substantial: 
with the extra time axis alignment, models can be trained to deal with fine-grained tasks, 
and predict temporal action localization and segmentation.

In this paper, we  tackle the sentence-to-video temporal alignment problem,
and propose a Temporal Alignment Network~(TAN) that ingests a video sequence and its associated narrative sentences, 
attends to a large temporal context in both, and is able to: 
(1) determine if a sentence is alignable with the video; 
and (2) if it is alignable, then determine its temporal alignment.
Given all the challenges described above, training such a network on raw instructional videos, 
{\em e.g.}, HowTo100M, is clearly a non-trivial task.
To this end, we propose a novel method for denoising, 
by co-training TAN with an auxiliary dual encoder network.
By design, these two networks use complementary architectures:
TAN iteratively attends to temporal context from both visual and textual modalities,
establishing accurate alignment for sentences that are alignable;
while the dual encoder processes visual and textual modalities independently, 
which enables it to spot unalignable sentences at ease, 
{\em e.g.}, sentences that emit low alignment score to all frames within the video.
The output from these two networks can be treated as two different views for alignment,
and their {\em mutual agreements} are adopted for co-training.

In addition to introducing the model and training methodology, 
we make the following contributions:
(1) We manually annotate an 80-video subset of HowTo100M, named \textbf{HTM-Align},
by assigning the visually related sentences to their corresponding timestamps and annotating visually unrelated ones.  
This aligned subset is used to evaluate the model's performance 
and is released publicly;
(2) We train the model on the HowTo100M dataset, 
and demonstrate a significant improvement in alignment over prior work (MIL-NCE approach of~\cite{Miech20} in particular);
(3) We apply the trained model in both the zero-shot and fine-tuned settings to multiple downstream video tasks and achieve state of the art results on both settings. 
This includes text-video retrieval on YouCook2~\cite{Zhou18} and 
weakly supervised video action segmentation on Breakfast-Action~\cite{Kuehne12};
(4) We use the automatically-aligned HowTo100M annotations to finetune
the \emph{backbone model}, and observe improved performance 
on downstream action classification tasks.

\vspace{-2mm}
\section{Related Work}
\vspace{-1mm}
\noindent \textbf{Joint Visual-Textual Learning}
has a long history in computer vision.
As examples, early work from Mori et al.~\cite{Mori99}
explored the connection between image and words in paired text documents, 
and~\cite{Weston11} learnt a joint image-text embedding for the case of class name annotations.
Recent works like CLIP~\cite{Radford21} and ALIGN~\cite{Jia21} show that 
large-scale paired image-caption data combined with a simple noise contrastive learning loss 
is able to learn a powerful visual representation.
In video domains, this is also true, as shown by MIL-NCE~\cite{Miech20}, ALBEF~\cite{Li21ALBF}, 
and VideoClip~\cite{Xu2021videoclip}. \\[-8pt]

\par{\noindent \textbf{Visual-Textual Retrieval}}
learns a joint embedding space for both vision and language,
either using a dual encoder~\cite{Bain21,Gong14a,Gong14b,Klein15,Pan16,Plummer17,Dong19,Miech19,Radford21,Jia21}, 
where visual and textual inputs are independently encoded, 
or a joint encoder, 
constructed with multimodal Transformers~\cite{Tan19_lxmert,Lu19_Vilbert,Su19_Vlbert,Lu20_12in1,Chen20_Uniter,Li20_Unicoder,Zhou20}, 
where vision and text inputs are fed into the cross-modal attention to compute the similarity. 
Despite being more accurate, the incurred computation of the joint encoder limits its use for large-scale retrieval systems.
In~\cite{Miech21}, 
the authors propose to speed up the process by only using the joint encoder for re-ranking.
In this work, we also use both joint and dual encoders, 
but for a different purpose -- to exploit their complementary information for co-training.\\[-8pt]

\par{\noindent \textbf{Visual-Textual Alignment}}
aims to temporally assign words or sentences to the corresponding
video segments.  A similar task is weakly-supervised action
segmentation that tries to delineate the video segments corresponding
to a given action
list~\cite{Bojanowski14,Bojanowski15,Kuehne17,Ding18,Huang16,Richard18,Chang19,Zhukov2019,Li19cdfl,Zhang21tqn}.
In transcript
alignment~\cite{Sankar06,Cour08,Sankar09,Zhu15,Tapaswi15}, where
instead of an action list, scripts describing a series of events in
the video are given, the goal is to assign each of the script texts to
the appropriate segment~(shot) of the video.  More closely related to
our goal are methods that seek a global alignment between sequences
with soft Dynamic Time Warping (DTW)~\cite{Cuturi17}. The recent
Drop-DTW~\cite{Dvornik21} proposes to handle outliers in the sequences
by allowing the alignment process to automatically skip certain
steps. This is similar to our aim of identifiying non-alignable
sentences. However, since in HowTo100M the order of the alignable sentences 
does not follow the original order of 
the subtitles, this rules out the use of DTW-type approaches.\\[-8pt]

\par{\noindent \textbf{Co-training and Self-training}}
are common techniques for unsupervised and weakly supervised learning.
Co-training~\cite{Blum98} builds two models 
to learn the different views of the data, 
while using one to expand the training set for the other.
It has recently been used for representation learning~\cite{Tian20,Han20coclr}.
Self-training refers to the process of training on 
pseudo-labels generated from a model's own predictions.
It has been used for image classification~\cite{Caron18,Xie20_student,Caron20,Asano20}, 
object detection~\cite{Cinbis15}, and machine translation~\cite{He20}. 
Our work is related to this line of research, 
where the TAN and the auxiliary network self-correct the noisy annotations, 
such that both networks can gradually improve their performance by training on cleaner data. \\[-8pt]

\vspace{-1mm}
\par{\noindent \textbf{Supervised Action Segmentation \& Detection}}
have been extensively studied on numerous video datasets, {\em e.g.}~Breakfast-Action~\cite{Kuehne12}, YouCook2~\cite{Zhou18}, Charades~\cite{Sigurdsson2016HollywoodIH}, ActivityNet~\cite{Caba15}, EPIC-Kitchens~\cite{Damen2020RESCALING}.
For segmentation, the goal is to densely classify each time point of the video into one of the pre-defined action categories~\cite{Kuehne12,Rohrbach12,Fathi13,Bhattacharya14,Bojanowski14,Singh16,Lea17,Lei18,Farha19,Chen20_seg}.
Research has focused on designing effective modules to 
capture dependencies between different video chunks~\cite{Singh16,Lea17,Lei18,Farha19}.
For detection, 
the goal is to localize the sparsely distributed action segments,
{\em i.e.}~annotation is non-contiguous.
In general, there are two-stage approaches that consist of a separate action proposal stage and a classification stage~\cite{Shou16,Xu17,Zhao17,Chao18,Lin18}, 
and one-stage approaches that combine both~\cite{Yeung16,Nawhal21}.

\begin{figure*}[!htb]
	\centering
	\includegraphics[width=0.95\textwidth]{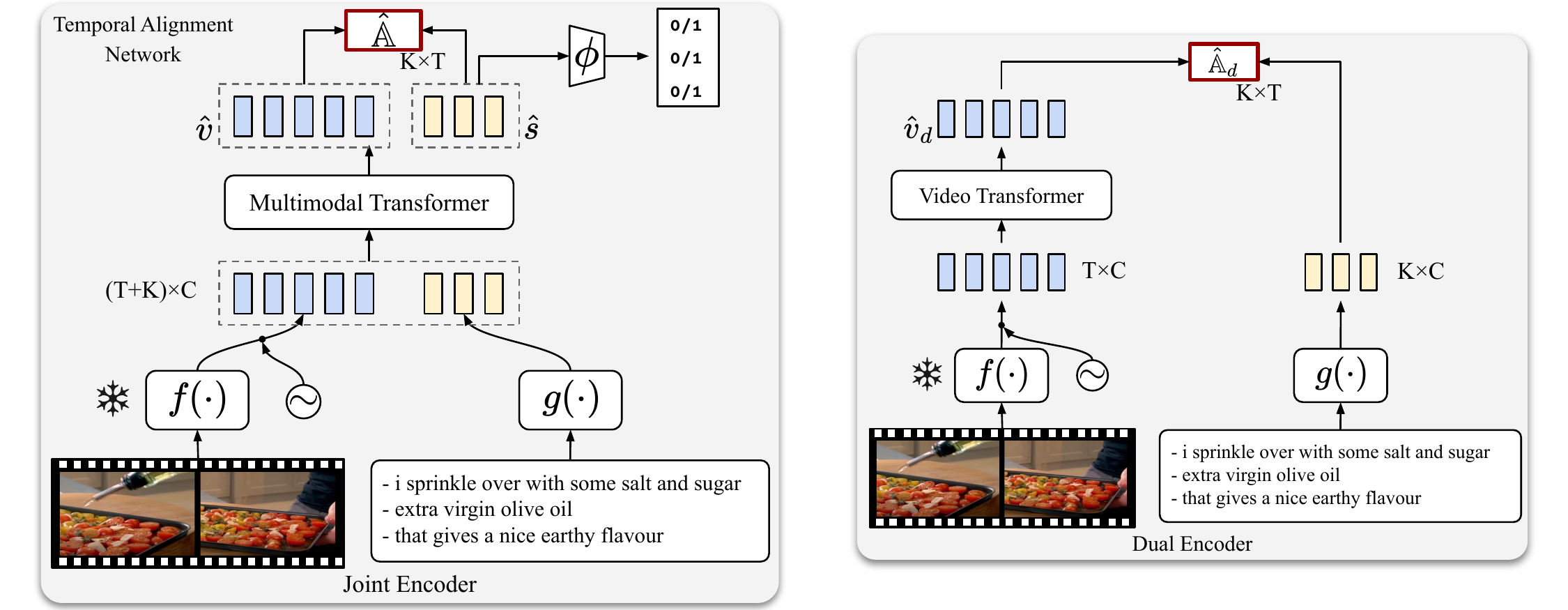}
	\vspace{-3mm}
	\caption{\footnotesize{{\em Left}: The Temporal Alignment Network (TAN) 
takes an untrimmed long video as input, 
and first extracts the visual and textual features by a pre-trained 3D ConvNet~($f(.)$)
and a pre-trained text module~($g(.)$).
The visual features and textual features are concatenated 
and passed into a Multimodal Transformer Encoder, 
a.k.a. \emph{joint encoder}, 
where the attention can capture the interaction between the visual and textual modalities.
A linear head $\phi$ classifies 
the alignability of the output text embedding.
{\em Right}: To train the TAN on noisy instructional videos,
we build an auxiliary  \emph{dual encoder},
which takes the same visual and textual features as input, 
but only use a Video Transformer Encoder to process the video data with self-attention.
For both TAN and the dual encoder, similarity matrices $\hat{\mathbb{A}},\hat{\mathbb{A}}_d$ are computed between the output text features and the output visual features respectively, which are used at the co-training stage, as introduced in Section~\ref{subsec:cotraining}.}}
\vspace{-16pt}
\label{main_fig}
\end{figure*}

\vspace{-1mm}
\section{Method}
\vspace{-1mm}
We start by describing the problem scenario in Sec~\ref{subsec:problem_scenario}, 
followed by the architecture for our proposed alignment network in Sec~\ref{subsec:tan}. 
In Sec~\ref{subsec:training}, we describe a na\"ive training procedure on raw instructional video,
with the text-video correspondence provided by YouTube ASR, despite the considerable noise.
In Sec~\ref{subsec:cotraining}, 
we present the co-training method, 
that exploits the {\em mutual agreement} between the alignment network 
and an auxiliary dual encoder,
and is able to simultaneously denoise and learn 
from the noisy narrated instructional videos.

\vspace{-1mm}
\subsection{Problem Scenario}
\vspace{-1mm}
\label{subsec:problem_scenario}
Given an untrimmed instructional video $\mathcal{X} = \{\mathcal{I}, \mathcal{S}\}$, 
where $\mathcal{I} = \{I_1, I_2, \dots, I_T\}$ refers to the corresponding video with $T$ frames, 
and $\mathcal{S} = \{S_1, \dots, S_K\}$ 
denotes the $K$ given sentences~(ordered by time). 
For each sentence, we also have their timestamps obtained from YouTube ASR 
({\em e.g.} $[t_k^{\text{start}}, t_k^{\text{end}}]$ for the $k$-th sentence).
In this paper, our goal is to train a temporal alignment network on a video dataset of instructional videos,
which takes the videos and sentences as inputs, and outputs a textual-visual similarity matrix~($\hat{\mathbb{A}}$),
as well as an alignability score for each sentence:
\vspace{-3mm}
\begin{align}
    \{ \hat{y}, \hat{\mathbb{A}} \} = \Phi(\mathcal{X}; \Theta),\quad \hat{\mathbb{A}} \in \mathbb{R}^{K \times T}
\end{align}
where $\hat{y} \in \mathbb{R}^{K \times 2}$ refers to binary scores for all sentences, 
indicating whether the sentence is alignable.
$\hat{\mathbb{A}}$ denotes the similarity matrix between frames and the given sentences, 
where for any alignable sentence it should emit a higher score with its corresponding video timestamps than others,
and $\Theta$ are the parameters of the model.

\vspace{1mm}
\subsection{Temporal Alignment Network~(TAN)}
\label{subsec:tan}
As shown in Figure~\ref{main_fig}~(left), 
the alignment network takes a video sequence and its associated narration / text sentences as input,
and attends to the long temporal contexts in both, in order to:
(i) determine if a sentence is alignable with the video~($\hat{y}$), and
(ii) output the alignment matrix~($\hat{\mathbb{A}}$).
Next, we describe the alignment network, consisting of a visual-textual backbone, 
Multimodal Transformer, and alignability prediction module.\\[-6pt]

\noindent \textbf{Visual-Textual Backbone. }
Given a long instructional video ({\em e.g.} 64s) with its associated sentences,
we first extract the visual and textual features with pre-trained networks.
Specifically, based on MIL-NCE~\cite{Miech20},
we use their pre-trained S3D-G backbone to extract video features,
and a 2-layer MLP with the word2vec embeddings~\cite{Mikolov13a} to extract sentence features.
\vspace{-1mm}
\begin{align}
    v = f(\mathcal{I}) \in \mathbb{R}^{T\times C}\quad\quad
    s = g(\mathcal{S}) \in \mathbb{R}^{K\times C}
\end{align}
$v, s$ refer to the computed video and text features respectively,
and each is of dimension $C$, in general, $T \gg K$.\\[-6pt]

\vspace{-1mm}
\noindent {\bf Multimodal Transformer.}
This module jointly processes the visual-textual features~($v, s$) with a multi-layer Transformer Encoder, 
which iteratively attends to both modalities to establish the text-to-video correspondence:
\vspace{-1mm}
\begin{align}
    [\hat{v}; \hat{s}] = \Phi_{\textsc{MT}}([v+\textsc{TE}; s])
\end{align}
where $\Phi_{\textsc{MT}}$ refers to the Multimodal Transformer Encoder,
\textsc{TE} denotes the learnable temporal embedding,
$\hat{v}\in \mathbb{R}^{T\times C}$ and $\hat{s}\in \mathbb{R}^{K \times C}$ are 
the output visual and textual embeddings from the Multimodal Transformer, and the
``$[;]$'' symbol denotes concatenation.
The alignment matrix $\hat{\mathbb{A}} \in \mathbb{R}^{K \times T}$ is computed via cosine similarity:
\vspace{-1mm}
\begin{align}
    \hat{\mathbb{A}}_{[i,j]} = \frac{\hat{s}_i \cdot \hat{v}_j}{\|\hat{s}_i\| \|\hat{v}_j\|} 
\end{align}

\noindent {\bf Alignability Prediction Module. }
Apart from estimating the alignment matrix, 
another main functionality of the alignment network is to infer whether a particular sentence is alignable or not. 
This is achieved by training a single linear layer~($\phi(\cdot)$) on the textual features, as shown in Figure~\ref{main_fig}~(left):
\vspace{-6pt}
\vspace{-2mm}
\begin{align}
    \hat{y} = \phi_{\text{align}}(\hat{s})
\end{align}
where $\hat{y} \in \mathbb{R}^{K \times 2}$ refers to the binary predictions for all sentences, 
deciding if the sentence is alignable or not.

\subsection{Training}
\vspace{-1mm}
\label{subsec:training}
In this section, 
we describe a na\"ive training procedure for the alignment network with contrastive learning, 
on the  instructional videos with YouTube ASR timestamps.
Note that, at this stage, all the sentences have their corresponding video timestamps, and are treated as alignable. 
Hence, the alignability prediction module can not be trained here. \\[-6pt]

\vspace{-2mm}
\noindent {\bf Temporal Correspondence. }
For a video with $K$ sentences,
we directly convert its YouTube ASR results into 1D binary masks,
with 1's at the timestamps where the sentence is being spoken by the demonstrator, {\em i.e.}, $\mathcal{Y} = \{ m_1, \dots,  m_K\}$, 
where $m_i \in \mathbb{R}^{1 \times T }$.
The objective is therefore to jointly optimize the visual-textual embedding, 
such that the similarity score between the sentence and its corresponding visual frames is maximised. 
The training objective is constructed as:

\vspace{-4mm}
{\footnotesize
\begin{align}
\mathcal{L}_{\textsc{TC}} = 
- \sum_{k=1}^{K}\log{
    \frac{\sum_{i \in \mathcal{P}_k}{\exp{(\hat{\mathbb{A}}_{[k,i]} / \tau)}}}
         {{\sum_{i \in \mathcal{P}_k} {\exp{(\hat{\mathbb{A}}_{[k,i]} / \tau)}}} + 
         {\sum_{j \in \mathcal{N}_k} {\exp{(\hat{\mathbb{A}}_{[k,j]} / \tau)}}}}
         }
\label{eq:tc_loss}
\end{align}}
\vspace{-4.0mm}

\noindent where $\mathcal{P}_k \in \{m_k = 1\}$, 
$\mathcal{N}_k \in \{m_k = 0\}$ refer to the sets 
consisting of positive and negative pairs, respectively. 
$\mathcal{L}_{\textsc{TC}}$ resembles a variant of the InfoNCE loss~\cite{Oord18}.\\[-8pt]

\noindent {\bf Discussion}. 
Given the groundtruth annotation for alignment,
optimizing $\mathcal{L}_{\textsc{TC}}$ would be trivial.
However, on raw instructional videos 
where the provided YouTube ASR timestamps are highly unreliable 
with an extremely high noise ratio,
na\"ively optimising $\mathcal{L}_{\textsc{TC}}$ leads to sub-optimal results, as will be demonstrated in Section~\ref{subsec:ablation}.

In general, the noise sources from the raw instructional videos 
can be mainly categorised into three types, as shown in Figure~\ref{fig:align}:
\emph{First}, 
the majority of the given sentences are actually not correlated to the video content (unalignable),
{\em e.g.}~greeting, chatting;
\emph{Second}, there is an alignment offset, 
in that the temporal interval of the spoken sentence rarely aligns with the video segments it refers to;
\emph{Third}, the demonstrator  often makes statements that do not follow the same order as their action, 
which rules out the use of DTW-type approaches.

\vspace{-1mm}
\subsection{Co-Training}
\vspace{-2mm}
\label{subsec:cotraining}
In this section, 
we propose a novel {\em co-training} method to both denoise the instructional videos and train the alignment network.
Specifically, we introduce a dual encoder~(Section~\ref{subsubsec:dual_enc}), 
which can be seen as a collaborator to the alignment network. 
This procedure is detailed below.

\begin{figure*}[t]
	\includegraphics[width=\textwidth]{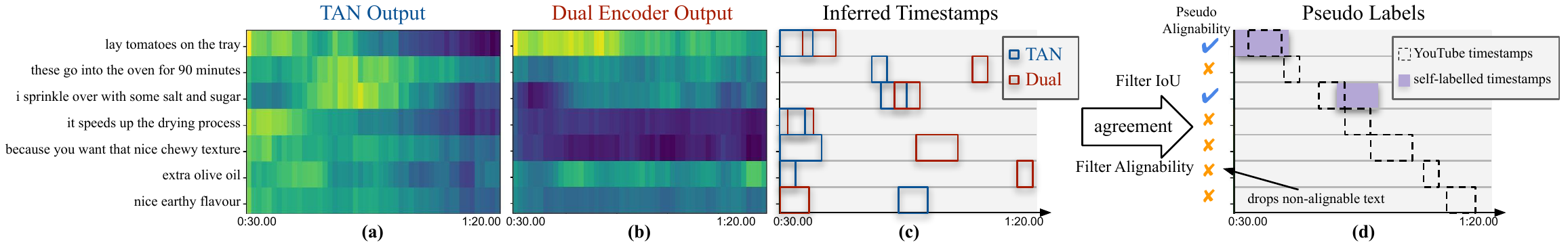}
	\vspace*{-8mm}
	\caption{\footnotesize{
    Illustration of denoising by mutual agreement.
    The video sample is the same as in Figure~\ref{fig:align}.
	\textbf{(a)}: The alignment matrix $\mathbb{A}$ from the TAN after Stage-1 training.
	\textbf{(b)}: The alignment matrix $\mathbb{A}_d$ from the dual encoder
	after Stage-1 training.
	\textbf{(c)}: The most alignable timestamps are inferred from both alignment matrices.
	\textbf{(d)}: By filtering the IoU of pseudo-timestamps 
	and filtering the alignable/non-alignable 
    text denoted by \textcolor{googleblue}{\textbf{\cmark}} / 
    \textcolor{googleorange}{\textbf{\xmark}},
	the model dynamically chooses aligned temporal segments to train,
	and ignores non-alignable ones.
	For this example, the self-labelling process corrects the timestamps of the 1st and 3rd sentences, 
	and marks the rest of the sentences as non-alignable.
	It roughly matches the human judgement of the alignment
	as shown in Figure~\ref{fig:align}.
	The alignment matrix values shown here are computed from the trained model-A in Table~\ref{table:ablation_loss}.}}
	\vspace{-6mm}.
	\label{fig:loss}
\end{figure*}

\vspace{-4mm}
\subsubsection{Dual Encoder}
\vspace{-2mm}
\label{subsubsec:dual_enc}
As shown in Figure~\ref{main_fig}~(right),
the dual encoder independently processes the visual features with a Transformer Encoder~\cite{Vaswani17}.
It is designed to be complementary to the alignment network:
for example, the dual encoder is fast and lightweight, 
which enables training on a large number of visual-text pairs,
however, it only allows both modalities to communicate at the end, 
hence is unable to capture the textual contexts, 
and it is more sensitive to detect unaligned texts;
while the proposed TAN, consisting of a Multimodal Transformer,
always has access to both modalities, and can learn to establish visual-textual correspondence within the network. 
Despite being beneficial for the temporal alignment task,
the TAN  is slow and computationally demanding, limiting its ability for contrasting with large-scale and diverse negative visual-textual pairs. 
Formally, for the dual encoder, we have: 
\vspace{-3mm}
\begin{align}
    \hat{v}^{d} = \Phi_{\textsc{d}}(v + \textsc{TE}) \in \mathbb{R}^{T\times C}
\end{align}
\vspace{-4mm}

\noindent where $\Phi_{\textsc{d}}$ refers to the Video Transformer, 
and $\textsc{TE}$ is the learnable temporal embedding for supplying the temporal ordering.
A textual-visual cosine similarity matrix $\hat{\mathbb{A}}_d\in \mathbb{R}^{K \times T}$ 
from the dual encoder is computed as:
\vspace{-3mm}
\begin{align}
    {\hat{\mathbb{A}}}_{d [i,j]}  = \frac{s_i \cdot \hat{v}^{d}_{j}}{\|s_i\| \|\hat{v}^{d}_{j}\|}
\end{align}
\vspace{-9mm}

\subsubsection{Denoising by Mutual Agreement}\label{subsubsec:denoising}
\vspace{-1mm}
To denoise the YouTube ASR annotations,
we generate pseudo-labels (both the alignability and the timestamps) 
by verifying the \textit{mutual agreement} between the output alignment matrices 
from the alignment network and dual encoder, {\em i.e.}~$\hat{\mathbb{A}}$ and $\hat{\mathbb{A}}_d$.
The verification process is executed in three steps: \\[-4pt]

\vspace{-2mm}
\noindent \textbf{(a) Infer Timestamps. }
During training, for each sentence,
we use the two output alignment matrices $\hat{\mathbb{A}},\hat{\mathbb{A}}_d \in \mathbb{R}^{K \times T}$ (Figure~\ref{fig:loss}-a,b) to infer the most plausible aligned timestamps.
To avoid outlier points, for the $k$-th sentence, 
we scan its corresponding similarity row by averaging the scores within a temporal window, 
this window is of the same size as its original YouTube timestamp label, {\em i.e}~sentence by the demonstrator.

Afterwards, 
we pick the most confident prediction by taking the $\argmax$.
Note that, 
such operation ends up with a single temporal window
with the same duration as the YouTube timestamp.
That is to say, 
we only shift the temporal position of the original YouTube label to its most confident prediction.
At this step, for the $k$-th sentence, 
we obtain two `shifted' timestamps $\hat{m}_k$ and $\hat{m}^d_k$,
one from the alignment network, 
the other from the dual encoder respectively, 
as shown in Figure~\ref{fig:loss}-c.  \\[-4pt]

\vspace{-2mm}
\noindent \textbf{(b) Alignment overlap using IoU. }
Given the inferred timestamps for sentence $k$,
we compute an Intersection-over-Union (IoU) score to measure 
the agreement between the shifted timestamps: 
\vspace{-4mm}
\begin{align}
\text{IoU-score}_k = \frac
{\hat{m}_{k} \cap \hat{m}^d_{k}}
{\hat{m}_{k} \cup \hat{m}^d_{k}}
\end{align}
\vspace{-5mm}

\noindent 
A high IoU score indicates the sentence is very likely to be 
aligned with the inferred timestamps.
For a batch, we filter the sentences with a positive IoU score,
and update their timestamps by the union of their inferred timestamps $\hat{m}_{k} \cup \hat{m}^d_{k}$. 
Empirically, we find this operation roughly updates the timestamps for about 30\% of the sentences.
For the sentences with zero IoU score, 
we keep their YouTube timestamps unchanged.
Such an operation ends up with a set of updated timestamps 
$\{\hat{m}'_1, \dots, \hat{m}'_K\}$ for all sentences.

In addition, to reflect each sentence's alignability, 
we can compute an average cosine similarity score falling into the new temporal segment. Formally, for the $k$-th sentence,
\vspace{-2mm}
\begin{align}
\epsilon_k = \frac{1}{\sum{\hat{m}'_k}}
                       \sum{\hat{m}'_k\cdot (\hat{\mathbb{A}}+\hat{\mathbb{A}}_d)_{[k,:]}} 
\end{align}
\vspace{-4mm}

\noindent where $\epsilon_k$ refers to the alignment score.
To put it simply, if a sentence has positive IoU-score,
we compute its align-score within the union of inferred timestamps;
if it has zero IoU-score, we compute its align-score within its original YouTube timestamps.\\[-4pt]

\vspace{-2mm}
\noindent \textbf{(c) Filter Alignability. }
To filter the alignability scores,
{\em i.e.}, $\{\epsilon_1, \dots, \epsilon_K\}$,
we introduce a hyper-parameter $\alpha \in [0,1]$,
within a sample batch, we treat the sentences with the top $100\alpha\%$ of align-score as positive, 
and the bottom $100(1-\alpha)$\% sentences as negative.
This gives binary pseudo-labels for alignability, denoted as $y_{\text{pseudo}}$.
The alignability prediction module can thus be trained for binary classification
with a cross-entropy loss~(as shown in Figure~\ref{main_fig}), 
{\em i.e.}, $\hat{\mathcal{L}}_{\text{Alignability}} = \text{CE}(\hat{y}, y_{\text{pseudo}})$.

Intuitively, this is to say, 
a sentence is treated as being alignable 
if both the alignment network and the dual encoder agree the sentence 
has a high similarity with its corresponding time stamps.
Also the $\mathcal{L}_\text{TC}$ (Equation~\ref{eq:tc_loss})
is only trained for the top $100\alpha\%$ of the sentences. 
In our experiments, we sweep $\alpha\in\{0.25, 0.5, 0.75\}$.

\vspace{-10pt}
\subsubsection{Training Cycle}
\vspace{-2mm}
To summarize, the training can be divided into two stages. At the first stage~({S1: Initialization}), 
both the alignment network and dual encoder are trained with ${\mathcal{L}}_{\text{TC}}$ using the given YouTube timestamps as labels.
Once warmed up, 
the new pseudo-labels will be generated from the mutual agreement 
between alignment network and dual encoder \emph{on the fly}, 
and starts the second stage training~({S2: Co-Training}),
with $\mathcal{L}_{\text{total}} = \hat{\mathcal{L}}_{\text{TC}} +
\hat{\mathcal{L}}_{\text{Alignability}}$.
Note that it is not necessary to iterate S1 and S2,
because in S2 the quality of pseudo-labels can be improved 
along the training with an EMA mechanism (introduced next).
By default in our experiments,
we train S1 for 50k iterations, 
and train S2 for another 50k iterations.
It accounts for 8 epochs on HTM-370K per stage.

\vspace{-12pt}
\subsubsection{ \noindent \textbf{Self-labelling with EMA }}
\vspace{-2mm}
Na\"ively using the {\em mutual agreement}
between the alignment network and dual encoder
for co-training can lead to trivial solutions,
where the alignment network and the dual encoder learn to ``collaborate'' with each other
and assign high similarity scores to certain fixed timestamps.
We avoid this `collapse problem'  by keeping an Exponential Moving Average (EMA) of the model 
similar to BYOL~\cite{Grill2020byol}.
The EMA branch is only slowly updated and used to generate the agreements for denoising 
as introduced in Section~\ref{subsubsec:denoising}. 
The main branch is trained with the updated timestamps and alignability.
We use the same momentum coefficient as that from BYOL in our experiments ($0.99$).
By default, all evaluations use the main branch. 
\vspace{-2mm}
\section{Experiments}
\vspace{-1mm}
In this paper, 
we train the proposed temporal alignment network on a subset of the HowTo100M dataset~\cite{Miech19}.
To start, we first describe the data preparation process,
and present the annotated visual-textual aligned subset of HowTo100M~(named \textbf{HTM-Align}) for evaluation.
Then we describe the implementation details and ablation studies for the alignment task. 

\vspace{-1mm}
\subsection{Data Preparation}
\vspace{-1mm}
HowTo100M is a large-scale instructional video dataset crawled from YouTube,
consisting of around $1.2$M videos and their generated text transcripts from speech~(ASR).
The start-end timestamps of each sentence are provided by ASR, 
but they are often {\em not semantically aligned} with the visual scene~(Figure~\ref{fig:align}). 

\vspace{-.4cm}
\subsubsection{HTM-370K~(Training)} 
\vspace{-2mm}
We mostly use a subset of the original HowTo100M for training,
with 370K videos from the `Food \& Entertaining' categories,
consisting of 32\% of the videos of the entire HowTo100M dataset.
Apart from the mis-alignment issue, we also find three other issues in the subtitles:
incorrect language translation, duplicated text, and incomplete sentence fractions.
As dataset pre-processing, 
we conduct an automatic curation with open-sourced BERT-based model.
The full details of automatic curation can be found in the~\app.

After automatically processing and filtering out low-quality subtitles, 
we end up with a subset of 370K instructional videos, thus the name \textbf{HTM-370K}.
Note that all the cleaning steps are automatic, using models trained with self-supervised learning.
We attribute the pre-processing of HowTo100M as a small contribution,
and we will make all cleaned video IDs and subtitles publicly available.

\vspace{-7pt}
\subsubsection{HTM-Align~(Evaluation)} 
\vspace{-2mm}
We randomly pick 80 videos from the HTM-370K as a hold-out testing set for evaluation purpose.
These videos range from $2$ to $16$ minutes, totalling 10 hours.
We manually label the alignability for each sentence, 
{\em i.e.}~binary annotation.
For those alignable ones, 
we further align them to the video segments with start-end timestamps.
In total, 
49K sentences are manually examined, with 13K of them being manually aligned.
On average each video contains 61 sentences, and 17 of them are visually aligned.

Unlike the existing YouCook2 benchmark, 
where annotators only rephrase fixed recipe steps as the action description,
\textbf{HTM-Align} includes random instructional videos without a fixed recipe,
and are adopted from the demonstrators' narration with minor modification,
hence containing large diversity on both videos and texts.
The details of the annotation and examples can be found in the \app. 

\vspace{-2mm}
\subsection{Implementation Details}
\vspace{-2mm}
\label{sec:imp}
During training, 
we adopt a pre-trained S3D~(released by~\cite{Miech20}) as the video encoder.
Specifically, 
the S3D network outputs a single feature vector~(1024D) for every 16 frames, 
when the videos are decoded with 16fps, 
this accounts for 1 feature per second without temporal overlap.
For the text encoder, we use Bag-of-word~(BoW) based on Word2Vec embeddings.
By default, in each video we randomly sample a temporal window of 64 seconds 
(which is 64 continuous visual features, we also tried 32s and 128s in ablation study), 
and the corresponding subtitles within this window.
We train the model with AdamW optimizer and $10^{-4}$ learning rate, 
with a batch size of 64 videos.
Full implementation details are in the~\app. \\[-8pt]

\begin{table*}[!htb]
\footnotesize
\setlength\tabcolsep{5pt}
\centering
\begin{tabular}{cccc|cc|c|cc}
\toprule        
& \multicolumn{3}{c|}{\bf Basic Setting} & \multicolumn{2}{c|}{\bf Training Stages} 
& \multicolumn{1}{c|}{\bf Stage2 Settings} & \multicolumn{2}{c}{\bf Aligned-HTM}  \\
\midrule

model &  dataset & length~(\# sec) &\# tfm layers & S1:Init & S2:Self & Threshold $\alpha$  
& R@1$\uparrow$  & ROC-AUC$\uparrow$ \\ \midrule

{CLIP~(ViT-B/32)~\cite{Radford21}} & YFCC-400M  & --  & --  & --  & --  & --   & 16.8   & 71.7$^*$   \\ 
MIL-NCE~\cite{Miech20}  & HTM-Full (uncurated) & --   & --  & --     & --     & --   & 31.3  & 73.1$^*$   \\
\midrule
A  & HTM-370K & 64   & 6-6 &\cmark  & \xmark & --   & 45.8 & 73.0$^*$  \\
B  & HTM-370K & 64   & 3-3 &\cmark  & \xmark & --   & 42.3 & 72.6$^*$  \\
\midrule

C & HTM-370K & 64   & 6-6 & \cmark & \cmark & 0.25  & 42.5 & 79.7 \\ 
D & HTM-370K & 64   & 6-6 & \cmark & \cmark & 0.5   & {\bf 49.4} & 82.4 \\ 
E & HTM-370K & 64   & 6-6 & \cmark & \cmark & 0.75  & 48.8 & 82.2 \\ 
\midrule

F & HTM-370K & 32   & 6-6 & \cmark & \cmark & 0.5   & 41.1 & 77.5 \\ 
G & HTM-370K & 128  & 6-6 & \cmark & \cmark & 0.5   & 48.4 & 81.8 \\
H  & HTM-Full & 64   & 6-6 &\cmark  & \cmark & 0.5   & 49.2 & \textbf{82.6} \\
\bottomrule

\end{tabular}
\vspace{-8pt}
\captionof{table}{\footnotesize \textbf{Alignment results on the HTM-Align dataset.}
         $^*$: since the model does not have a binary classifier for alignability,
         for each sentence, we take its maximum logits over time as the alignability measurement to compute ROC-AUC.
         For the `\# tfm layers' column,
         we show the number of transformer encoder layers we use for the TAN and the dual encoder.}
\label{table:ablation_loss}\vspace{-.6cm}
\end{table*}

\vspace{-3mm}
\section{Alignment Results}
\vspace{-2mm}
In this section, 
we report the experimental results for our proposed temporal visual-textual alignment task on the \textbf{HTM-Align} dataset.
In detail, during inference, 
given the video with a sequence of sentences by the demonstrator,
we take the alignment matrix from our alignment network, $\hat{\mathbb{A}} \in \mathbb{R}^{K \times T}$,
with $K, T$ indicating the number of sentences and video timestamps respectively.

\vspace{-2mm}
\subsection{Metrics}
\vspace{-1mm}
We measure two metrics for the alignment task, Recall@1 and ROC-AUC value.
The \textbf{Recall@1} metric is a `pointing game' as introduced in~\cite{Zhukov2019}.
Specifically, for a considered sentence, 
if its maximally matched video timestamp falls into the groundtruth segment,
it is counted as being successfully recalled. 
The recall scores are then averaged across all the text segments.
The alignability prediction is a binary classification problem as introduced in section~\ref{subsec:problem_scenario},
we use ROC curve and report the \textbf{Area-Under-the-Curve} value (ROC-AUC).

\vspace{-2mm}
\subsection{Ablation Study}
\vspace{-1mm}
\label{subsec:ablation}
In this section, we investigate the effects of multiple design choices and discuss the results.\\[-8pt]

\vspace{-1mm}
\noindent {\bf Comparing with baseline.}
In Table~\ref{table:ablation_loss}, 
the first two rows are the baselines from CLIP~(ViT-B/32)~\cite{Radford21} and MIL-NCE~\cite{Miech20}.
Specifically, we compute the alignment similarity matrix using their textual and visual encoders, 
normalize the score following their pretrained paradigms, and compute the R@1 on top of the alignment matrix. 
Note that for ROC-AUC, since CLIP and MIL-NCE do not have a specific binary classifier,
for each text, we directly use its maximum similarity score (across the time axis) as an indicator of alignability.
First, CLIP~\cite{Radford21} is performing significantly worse than others on this alignment task.
A possible reason is that CLIP has only been trained on images, thus lacks video dynamics.
MIL-NCE is a strong baseline which has short-term temporal modelling (up to 3.2s) 
and was trained end-to-end on HowTo100M.
In our model-A, we take the pre-extracted visual and textual feature from MIL-NCE,
and train the transformers on the HTM-370K dataset
to learn a longer temporal context ({\em e.g.} 64s) for the alignment task.
Our result (model-A 45.8 R@1 vs MIL-NCE 31.3 R@1) shows that longer temporal context is useful for this alignment task. \\[-8pt]

\vspace{-1mm}
\noindent {\bf Effect of Transformer Depth.}
For both the alignment network and dual encoder, we use 6-layer transformers by default, 
as a balance between performance and training cost.
In model-B we also tried using 3-layer transformer and found it performs worse than 6-layer transformer (model-B vs A).
Using more than 6 layers takes more memory and sacrifices batch size. \\[-9pt]

\vspace{-1mm}
\noindent {\bf Effect of Co-Training.}
In the model-\{D,E\}, we apply the Stage-2 training (co-training) based on the model-A.
We observe that co-training brings a clear performance gain for the alignment task (model-\{D,E\} vs.~model-A,
3-4\% boost on R@1), confirming the effectiveness of the denoising procedure.
Note that model-C does not perform well due to the choice of alignability threshold $\alpha$, explained next.
\\[-9pt]

\vspace{-4mm}
\noindent {\bf Effect of Alignability Thresholds.}
For the choice of alignability threshold $\alpha$ (as introduced in Section~\ref{subsec:cotraining}),
which reflects a balance of data noise and diversity in the co-training procedure,
our model-\{C,D,E\} show $\alpha=0.5$ and $\alpha=0.75$ work similarly well for alignment metrics and 
$\alpha=0.5$ is slightly better for the R@1 metric.
However $\alpha=0.25$ leads to much worse performance.
We conjecture that a low value of $\alpha$ limits the diversity while training $\mathcal{L}_{\text{TC}}$ 
({\em i.e.}~$\mathcal{L}_{\text{TC}}$ learns from only 25\% of the sentences).\\[-9pt]


\vspace{-1mm}
\noindent {\bf Effect of Training Data.}
In model-H, we train the co-training stage on the automatically curated HTM-Full dataset, 
which includes all other non-cooking categories from HowTo100M comparing with HTM-370K. 
Comparing model-H with D on the alignment task,
 adding out-of-domain videos does not harm the alignment performance on our curated subset. \\[-9pt]


\vspace{-1mm}
\noindent {\bf Effect of Input Video Length.}
In Table~\ref{table:ablation_loss}, 
we vary the length of the input video to show if our alignment network benefits from the longer video context.
Indeed, the alignment network gets better performance when increasing the input video length from 32 to 64 seconds~(model-D vs model-F). 
We conjecture that sampling longer input video introduces more \emph{alignable} sentences, 
helps to reduce the temporal ambiguity for other sentences.
However, further increasing the input video length to 128 seconds 
gives a similar alignment performance~(model-G vs model-D),
we conjecture this is due the reduced batch size in training,
and the far-away visual context~({\em i.e.}\ 2 minutes or further) is less relevant for aligning the sentence.

\begin{table*}[!htb]
\begin{minipage}[t]{0.47\textwidth}
\centering
\footnotesize
\hspace{-6pt}
\begin{tabular}[t]{c|c|lll}
    \hline
    Method   & Trained on BF         & F-Acc$\uparrow$ & IoU$\uparrow$  & IoD$\uparrow$  \\ \hline
    MIL-NCE~\cite{Miech20}           & \xmark~(ZS) & 59.3  & 46.8 & 65.1 \\ 
    \textbf{Ours (S1+S2)}     & \xmark~(ZS)    & \textbf{65.1}  & \textbf{50.6} & \textbf{68.6} \\ \hline
    D$^3$TW~\cite{Chang19}  & \cmark & 57.0  & -    & 56.3 \\
    CDFL~\cite{Li19cdfl}    & \cmark & 63.0  & 45.8 & 63.9 \\
    DP-DTW~\cite{Chang21}   & \cmark & 67.7  & 50.8 & 66.5 \\ 
    \textbf{Ours (S1+S2)}     & \cmark                & \textbf{68.3}  & \textbf{51.7} & \textbf{69.3} \\ \hline
\end{tabular}
	\vspace{-5pt}
\caption{
\footnotesize \textbf{Temporal alignment on the Breakfast-Action (BF) dataset.}
We split the previous methods into two groups. For the upper group, 
the model has not seen any samples in Breakfast-Action dataset since Breakfast-Action videos are not download-able from YouTube.
For the lower group, the model is trained with weak supervision on the Breakfast-Action training set.}
\label{table:align}
\end{minipage}
\hfill
\begin{minipage}[t]{0.51\textwidth}
\footnotesize
\setlength\tabcolsep{4pt}
\centering
	\begin{tabular}[t]{c|c|cccc}
		\hline
		Method    & Trained on YC2  & R@1$\uparrow$  & R@5$\uparrow$  & R@10$\uparrow$ & Median R$\downarrow$ \\ \hline
		ActBERT~\cite{Zhu20}     & \xmark~(ZS) & 9.6  & 26.7 & 38.0 & 19       \\
		MIL-NCE~\cite{Miech20}   & \xmark~(ZS) & 15.1 & 38.0 & 51.2 & 10       \\
		MIL-NCE~\cite{Miech20}$^{\dagger}$     & \xmark~(ZS)& 13.9 & 36.3 & 48.9 & 11 \\
		TaCo~\cite{Yang21taco}   & \xmark~(ZS) & 19.9  & 43.2  & 55.7  & 8.0 \\
		\hline
		{Ours (S1)}         & \xmark~(ZS) & 16.8           & 41.3  & 54.8  & 8.0 \\ 
		\textbf{Ours (S1 + S2)}    & \xmark~(ZS) & \textbf{20.1}           & \textbf{45.5}  & \textbf{59.5}  & \textbf{7.0}  \\
		
		\hline
	\end{tabular}
	\vspace{-5pt}
\caption{\footnotesize \textbf{Text-based video retrieval on the YouCook2 (YC2) dataset.} 
ZS refers to ``zero-shot'', where the alignment network is only trained on HTM-180K, and directly evaluated on YouCook2. 
$\dagger$: reproduced in~\cite{yc2_leaderboard}.
For our results, \textbf{S1} denotes only training Stage-1 (initialization),
which is the model-A from Table~\ref{table:ablation_loss}.
\textbf{S1+S2} denotes training with two stages (initialization followed by co-training),
which is the model-E from Table~\ref{table:ablation_loss}.
}
\label{table:tv_retrieval}
\end{minipage}
\vspace{-10pt}
\end{table*}

\vspace{-2mm}
\section{Downstream Tasks}\label{sec:downstream}
\vspace{-2mm}
Apart from evaluating the alignment task on \textbf{HTM-Align},
we also test our alignment network on other downstream tasks.
Specifically, we evaluate on temporal action alignment~(using the alignment network)
and text-based video retrieval~(using the dual encoder due to speed considerations~\cite{Miech21}).
We also evaluate linear action classification on the backbone feature to show the effect of auto-aligned dataset. 
See the~\app~for full details.\\[-8pt]

\vspace{-3mm}
\noindent {\bf Datasets.}
To evaluate the alignment network, 
we use \textit{Breakfast-Action}~\cite{Kuehne12} 
and \textit{YouCook2}~\cite{Zhou18} for downstream tasks.
To evaluate the end-to-end representation learning,
we use \textit{UCF101}~\cite{Soomro12}, \textit{HMDB51}~\cite{Kuehne11} and 
\textit{K400}~\cite{Kay17}.
\vspace{-0pt}

\vspace{2mm}
\noindent {\bf Temporal Alignment on Breakfast-Action. }
Given a video with multiple actions and the corresponding action descriptions, 
the model needs to densely label each video timestamp with one given text description,
often defined as weakly-supervised action segmentation by the community.
Following previous work~\cite{Ding18,Chang19,Li19cdfl,Chang21},
we report three metrics:
frame-wise accuracy (\textbf{F-Acc}),
segment-wise Intersection-over-Union (\textbf{IoU}) and Intersection-over-Detection (\textbf{IoD}).
Please refer to ~\app~for more details.

We evaluate our method with both the zero-shot and finetune settings.
In the former case, 
our alignment network was \emph{only} trained on HTM-370K, 
and directly evaluated on Breakfast;
while in the latter, 
we finetune our alignment network on Breakfast with a soft-DTW loss~\cite{Cuturi17} stacked on top of the output alignment matrix for 50 epochs. 
During inference, the alignment network takes as input a single video and the given list of action labels, ~\ie `crack egg', `fry egg', \emph{etc},
and outputs the alignment matrix $\mathbb{A}$, 
which is passed through a DTW, ending up with the action boundaries.

As shown in Table~\ref{table:align}, 
in the zero-shot setting, 
our proposed alignment network surpasses the strong baseline~(MIL-NCE) by a large margin on all metrics~($>3\%$), and even achieves comparable results to those supervised approaches.
After finetuning, we see a further performance boost, 
obtaining state-of-the-art results.  \\[-9pt]

\vspace{2mm}
\noindent {\bf Text-based Video Retrieval on YouCook2.}
We evaluate the model for text-based video retrieval on the YouCook2 dataset.
For this task, 
we pass each pre-cropped video segment through the \emph{dual encoder}, 
and take the visual features~($v_{\text{enc}}$) from the Video Transformer Encoder.
Also we pass the task description phrases into the dual encoder and take the word2vec features.
For each query text,
we rank the video segments based on cosine similarity among 3.5k candidates.
Following previous works~\cite{Miech19,Miech20}, 
we report retrieval Recall @\{1,5,10\} and Median Rank.

As shown in Table~\ref{table:tv_retrieval}, 
under the zero-shot setting, 
where the proposed alignment network was only trained on HTM-370K,
our model surpasses previous works by a clear margin,
especially on R@5, R@10 and Median R.
Importantly, the results show that the co-training stage substantially 
improves the performance of the \emph{dual encoder} (R@1 20.1 vs 16.8),
also our method surpasses the baseline method MIL-NCE 
by a large margin (R@1 20.1 vs 15.1). 

\vspace{2mm}
\noindent {\bf End-to-end Representation Learning.} 
The output of the Temporal Alignment Network 
can be used to clean-up (automatically align) long-video datasets.
We use model-H to automatically align the HTM dataset,
and finetune the S3D-word2vec backbone \textbf{end-to-end} with an {Info-NCE} loss
on the auto-aligned text-video pairs for only 10 epochs.
We evaluate the visual representation by linear probing on action classification,
and find the auto-aligned HTM timestamps benefits the end-to-end video representation.
We refer the readers for more details in the~\app.

\begin{table}[htb!]
\footnotesize
\centering
\vspace{-0mm}
\begin{tabu}[t]{c|c|ccc}
    \hline
    Settings & Backbone & UCF101 & HMDB51 & K400 \\\hline
    \rowfont{\color{gray}} 
    reported by~\cite{Miech20}  & S3D  &  82.7 & 53.1 & - \\
    reproduce of~\cite{Miech20}   & S3D  &  82.1 & 55.2 & 55.7 \\
    finetuned with TAN            & S3D  &  \textbf{83.2} & \textbf{56.7} & \textbf{56.2} \\\hline
\end{tabu}
\vspace{-2mm}
\caption{
\footnotesize \textbf{Linear-probing action classification performance.}
We evaluate the end-to-end trained visual representations
on UCF101, HMDB51 and K400 by linear probing (LP).
We show the reported LP results from~\cite{Miech20} (1st row),
our reproduction of LP results of the official S3D weights (2nd row),
and our finetuned S3D performance with auto-aligned HTM 
under the exact same setting (3rd row).
}\label{table:end_to_end}
\vspace{-3mm}
\end{table}

\vspace{-4pt}
\section{Conclusion}
\vspace{-2mm}
In summary, 
we introduce a temporal alignment network, with a co-training method for denoising the instructional video datasets.
To evaluate the alignment accuracy we introduce a new benchmark dataset with 10 hours of videos, 
with the narrations manually aligned to corresponding video timestamp.
When evaluating on \textbf{HTM-Align}, 
Breakfast-Action, YouCook2, 
under zero-shot or finetune settings,
our model achieves state of the art results, 
surpassing multiple strong baselines~(MIL-NCE, CLIP).
We also show the proposed method can clean-up (by improving the alignment) 
large-scale public datasets and further improve the visual-textual backbone representations.

\clearpage
\section*{Acknowledgement}
\vspace{-1mm}
This research is funded by 
EPSRC Programme Grant VisualAI EP/T028572/1, a
Royal Society Research Professorship RP$\backslash$R1$\backslash$191132,  and 
a Google-DeepMind Scholarship.
We thank Charig Yang, Guanqi Zhan and Chuhan Zhang for proof-reading.

\begingroup
{\small
\bibliographystyle{ieee_fullname}
\bibliography{bib/shortstrings,bib/vgg_local,bib/vgg_other}
}
\endgroup

\clearpage
\noindent{\LARGE \textbf{Appendix}}
\appendix



\section{Automatic Data Curation for HowTo100M}

\begin{figure*}[t]
    \centering
	\includegraphics[width=\textwidth]{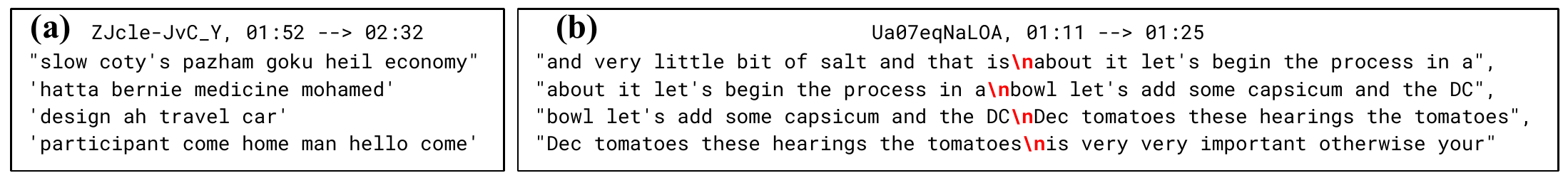}
	\vspace{-20pt}
	\caption{\small{Two problematic subtitle samples in HowTo100M dataset.
	Subtitles within a short temporal window are shown.
	\textbf{(a)} The presenter speaks in Thai, but the text is translated to English by ASR,
	resulting in nonsense text.
	\textbf{(b)} The linebreak ``\texttt{\textbackslash n}'' appears in every sentence segment, 
	resulting in repeated text segments.\vspace{5pt}} \label{fig:text}}
\end{figure*}

In this section, 
we describe the procedure for automatically curate HowTo100M dataset,
that targets three problems: Incorrect Language Translation,
Incorrect Linebreaks and Incorrect Sentence Partition.

\vspace{3pt}
\par{\noindent \textbf{Incorrect Language Translation}}
refers to the cases that 
the YouTube ASR system fails to detect the language and treats speech as English by default, 
thus generating incorrect text for non-English languages
(as shown in Figure~\ref{fig:text}-a, the original language of speech is Thai).
In order to detect such cases, 
we use open-sourced language detection library~\cite{nakatani2010langdetect}
which is a language classifier trained from Wikipedia that supports 53 languages.
The language detection library takes a phrase or sentence as input,
and returns a probability normalized over 53 languages. 
Specifically, for each HTM video, 
we randomly sample 5 subtitles, 
and average their probability of being English language, denoted as $\bar{p}_{\text{en}}$,
Then we discard the videos with $\bar{p}_{\text{en}} < 0.9$,
which accounts for about 3\% of the entire dataset.

\vspace{3pt}
\par{\noindent \textbf{Incorrect Linebreaks}}
refers to the cases where repetitive sentences exist in consecutive subtitles, 
as shown in Figure~\ref{fig:text}-b. We detect the linebreaks, 
and remove the duplicated sentence segments. 
In total, this operation corrects the captions for 68\% of the whole dataset.

\vspace{3pt}
\par{\noindent \textbf{Incorrect Sentence Partition}}
refers to the cases that the YouTube ASR system recognises unpunctuated sentences 
and wraps sentences to fit the width of the window, 
which generates incompleted sentence fractions.
To restore completed sentences, 
we first remove all the linebreaks 
and combine the sentence fractions into an unpunctuated paragraph,
then use an off-the-shelf BERT-based model to restore the punctuation~\cite{restore-punct21}
(the model is also trained without manual labelling),
where the completed sentences are cut out at the full stop.
We update the start and end timestamps for the completed sentence
by interpolating from the word timestamps.

\noindent Overall, the three automatic data curation steps filter out incorrect subtitles and improve the quality of subtitles.
Note that all the modules used in the curation are trained without human labelling, thus can be easily scalable.

\section{Details of HTM-Align Dataset}
In this section, we describe the annotation process, 
and show some examples from our \textbf{HTM-Align} dataset.

\begin{figure*}[!htb]
	\centering
	\includegraphics[width=0.9\textwidth]{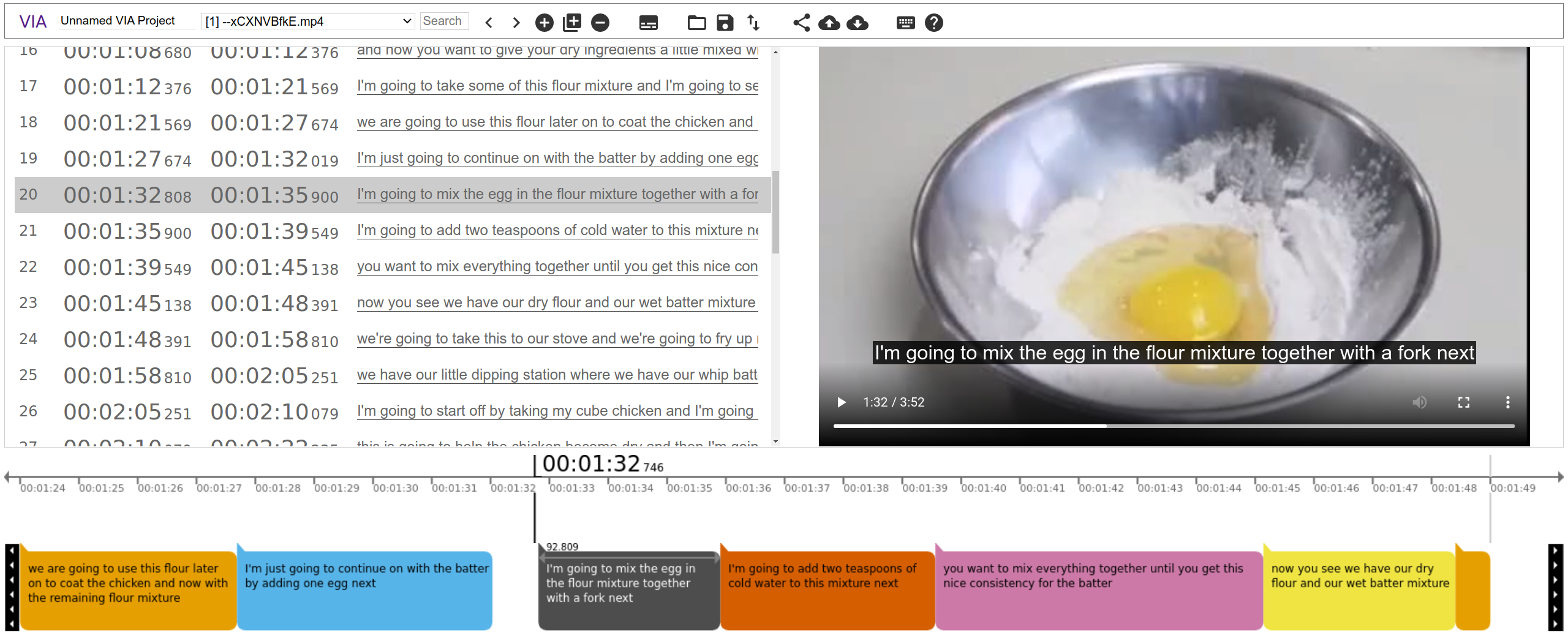}
	\vspace{-10pt}
	\caption{\small{A screenshot of the annotation process for the HTM-Align dataset.
	Two tasks are performed during annotation.
	The annotator (1) verifies the alignability of a subtitle sentence, 
	and if alignable, (2) adjusts the start-end timestamp to align subtitle sentence with the visual scene.
\vspace{-3pt}}}
\label{fig:via}
\end{figure*}

\subsection{Annotation Details}
In this paper, we use the open-sourced VIA~\cite{Dutta2016via} for annotation.
A screenshot of the annotation process is shown in Figure~\ref{fig:via}.
Given the video and its subtitles with start-end timestamps,
the annotator mainly performs two tasks: 
(1) determine if a subtitle sentence is alignable with the video~(i.e. visually related), 
(2) adjust the start-end timestamps to cover the visual content that is described by the ``alignable'' sentence.
Overall, the completed annotation contains two types of subtitles. 
For alignable subtitles, the sentence with its aligned start-end timestamp is provided,
for non-alignable subtitles, the sentence with its original start-end timestamp is simply inherited. 

\subsection{Examples}
\begin{figure*}[!htb]
	\centering
	\includegraphics[width=0.9\textwidth]{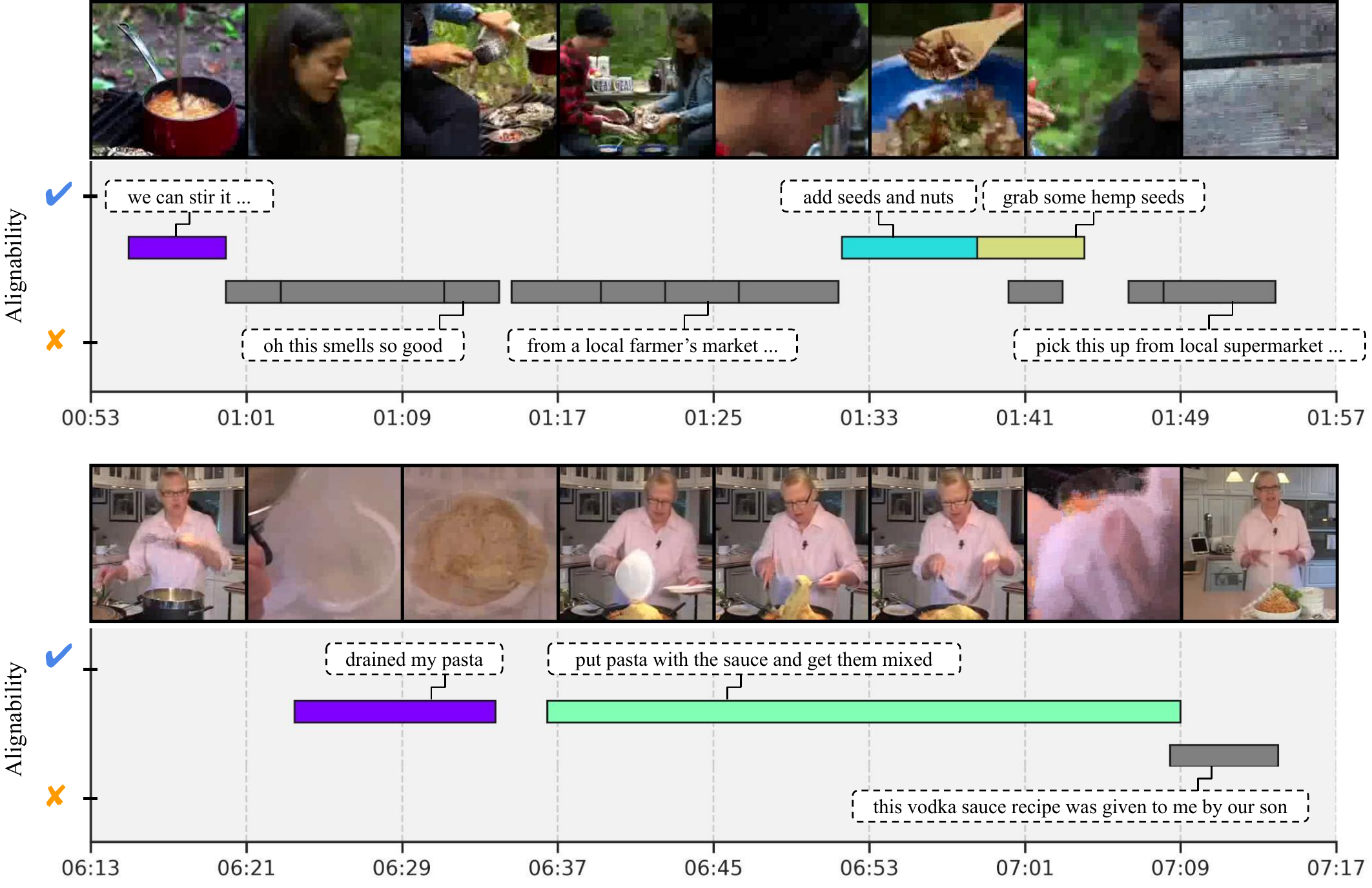}
	\vspace{-3pt}
	\caption{\small{Example videos from \textbf{Aligned-HTM} dataset.
	We visualize two 64-second video segments.
	The horizontal axis denotes the original video timestamp.
	\textcolor{googleblue}{\textbf{\cmark}} indicates the narrations with positive alignability,
	\textcolor{googleorange}{\textbf{\xmark}} refers to the texts that are visually NOT alignable,
	the width of the colored rectangle indicates the start-end timestamps of the text,
	which are all provided in the dataset. 
	For some of the non-alignable sentences we only show the rectangle but omit the text for clarity.
\vspace{-4pt}}}
\label{fig:aligned_htm_samples}
\end{figure*}

Here, we show two example videos with the aligned annotations 
from \textbf{HTM-Align} dataset in Figure~\ref{fig:aligned_htm_samples}.
More example videos are included in our \emph{project webpage}.
We show video on the top, 
the alignment annotations in the middle, 
and the TAN outputs at the bottom.
Note that the audio is copied from YouTube files only for presentation purpose,
in order to denote the raw timestamps of the YouTube subtitles.
Our model does not take audio input.

\section{Implementation Details}

\subsection{Architectural Details}

\vspace{0pt}
{\noindent \bf Visual Backbone.}
As introduced in paper Section~\ref{sec:imp}, 
we adopt a pre-trained S3D~\cite{Xie18s3d}~(pre-trained with MIL-NCE~\cite{Miech20}) as the video backbone.
Specifically, 
we decode the video with 16fps, feed S3D with 16-frame non-overlapping temporal window,
and take the feature vector~(1024D) just before the final fully-connected layer of S3D.
The original video is firstly resized to have shorter size equal to 256 pixels,
then the frames are center-cropped with $224\times 224$ resolution before feeding into the S3D.
The 1024D feature vectors are projected to 512D with a fully-connected layer before feeding into the transformer. 

\vspace{3pt}
{\noindent \bf Language Backbone.}
As mentioned in the main paper Section~\ref{sec:imp},
we adopt a Bag-of-word~(BoW) model based on Word2Vec embeddings.
Specifically, following~\cite{Miech20}, 
we use the Google-News self-supervised pre-trained Word2Vec embeddings (300D) from~\cite{Mikolov13a}. 
For each sentence, we take a maximum of 32 words and the word embedding is 
independently passed to two fully-connected layers and projected to $16\times 512\text{D}$,
and a max-pooling is applied to get a sentence embedding~(512D).

\vspace{3pt}
{\noindent \bf Transformer Modules.}
As shown in the main paper Figure~\ref{main_fig},
the TAN consists of a \emph{Multimodal Transformer}, while the auxiliary dual encoder contains a \emph{Video Transformer}.
For both of them, 
we use the pre-norm multi-layer transformer encoder implemented in~\cite{Radford21}.
As for the depth of transformer, we use a 6-layer transformer by default,
and we also ablate different depth (3-layer) in main paper Section~\ref{subsec:ablation}. 
The transformer uses 8-head attention mechanism, takes hidden vectors with 512D, 
and uses 2048D feed-forward dimensions.
For the video features, 
both transformers share a learnable position encoding to inject temporal information.
For the language features, we do not use positional encoding, as we found it tends to lead trivial solutions, 
where the model learns the temporal alignment \emph{only} by positional encoding.

\vspace{-0mm}
\subsection{Details for Downstream Datasets}

{\noindent \bf Breakfast-Action}~\cite{Kuehne12} 
includes 1.7k videos from a third-person-view for the breakfast preparation activities in kitchen.
There are in total 48 different actions,  plus a background action.
Here, action classes are only short phrases like `fry egg' and `pour milk',
and each video contains 6 action instances on average.

\vspace{5pt}
{\noindent \bf YouCook2}~\cite{Zhou18} contains 2k long videos covering 89 cooking recipes. 
The key procedures of the recipe are localized and annotated with sentences. 
On average, each video has 8 annotated segments, totalling 14K action segments.
We mainly report the text-video retrieval performance on about 3.5k validation segments.

\vspace{5pt}
{\noindent \bf UCF101}~\cite{Soomro12} contains 13k short (4-14 seconds) video clips spanning 101 human actions.

\vspace{5pt}
{\noindent \bf HMDB51}~\cite{Kuehne11} contains 7k short (1-6 seconds) video clips spanning 51 human actions.

\vspace{5pt}
{\noindent \bf K400}~\cite{Kay17} stands for Kinetics-400 dataset, which contains 
about 300K 10-second video clips for 400 human actions.

\vspace{1mm}
\subsection{Details for Downstream Tasks}

\paragraph{Temporal Alignment on Breakfast-Action.}
We evaluate this alignment task using TAN (without the auxiliary dual encoder), as explained in main paper Section~\ref{sec:downstream}. 
We first apply the same data pre-processing as in pre-training on HTM-370K,
~i.e.~decoding video with 16fps and extract S3D features with center-cropped 16-frame input, 
resulting in 1 feature per second without temporal overlap. 
For the language encoding, 
we use the same procedure and extract single vector~(512D) for each action class like `crack egg'.
Breakfast-Action dataset provides a background action class, 
which usually appears at the start and end of the video.
Note that we only take the non-background video segments 
(i.e. the middle part of video) 
to evaluate the temporal alignment task.
For a single video, 
the visual features with $T$ time steps and 
a list of $K$ text features are fed into the transformer module, 
resulting with the alignment matrix $\hat{\mathbb{A}}\in{\mathbb{R}^{K\times T}}$.
In practice, $T$ can be longer than 1 minute. 
To fit our positional encoding that is only trained for 64 seconds, 
we crop visual feature with a 64-second~(64 time stamps) sliding window,
and stitch multiple $\mathbb{R}^{K\times 64}$ alignment matrices as the final $\hat{\mathbb{A}}$.
The alignment matrix $\hat{\mathbb{A}}$ is then passed to 
Dynamic Time Warping~(DTW~\cite{Sakoe78}) to get the action boundaries.

{\noindent \bf Text-based Video Retrieval on YouCook2.}
The validation split of YouCook2 contains about 
3.5K paired video segments and action descriptions.
We evaluate this retrieval task with the auxiliary dual encoder,
as explained in main paper Section~\ref{sec:downstream}. 
For visual encoding,
we first use the same data pre-processing as in pre-training on HTM-370K,
{\em i.e.}~extracting S3D video features with 1 feature per second.
Following the prior works~\cite{Miech20,Yang21taco}, 
we pre-crop the original long video to obtain short video segments 
based on the annotated start-end timestamps for each annotated sentence.
Each video segment is fed into the video transformer of the \emph{dual encoder} 
to get the visual output~($\hat{v}_{d}\in \mathbb{R}^{{t\times 512}}$) 
where $t$ denotes the duration of the segment,
then a temporal average pooling is applied to get a 512D vector, representing each video segments. 
For textual encoding,
we extract the sentence feature~(a single 512D vector) for each action description,
such as `chop lettuce and place it in a bowl'.
Finally, each sentence embedding retrieves segment-wise video features 
by cosine similarity and the metric is calculated following~\cite{Miech20}.

\vspace{-0mm}
\section{End-to-end Representation Learning with auto-aligned HTM}
\vspace{-0mm}
\subsection{Details for End-to-end Training}
\vspace{-1mm}
We conduct the end-to-end trainings on the 25\% of all the original HowTo100M raw videos.
First, we use the model-H from the main paper Table~\ref{table:ablation_loss}
to compute the text-video alignment.
For each text, we store the timestamp with the highest similarity
as well as the alignability score.
For data sampling, we drop the texts with 
the lowest 50\% alignability score, 
and load the auto-aligned text-video pair for training.
To introduce more hard negatives for Info-NCE loss,
we randomly sample 2 text-video pairs from each long-video.
For each video clip, we follow~\cite{Miech20} and use 16 frames with 5 fps.
In this way, 
we fit a batch size of 32 text-video pairs (come from 16 source videos).
We load the official S3D-Word2Vec backbone from~\cite{Miech20},
and finetune the backbone on our auto-aligned HowTo100M subset for only 40 epochs 
-- equivalent to 10 epochs on full-set HowTo100M in terms of the number of iterations.
We use an AdamW optimizer with $10^{-4}$ learning rate and 
a cosine decay learning rate schedule for training.

\subsection{Details for Linear Probe Evaluation}
We keep the S3D backbone frozen, 
and only train a linear layer on the 1024-D visual feature
with a cross-entropy loss.
We use AdamW optimizer with $10^{-2}$ initial learning rate with
cosine decay learning rate schedule
and $10^{-3}$ weight decay to train the linear layer.
We both UCF101 and HMDB51, we use the split-1.
The network takes input as 16 frames video clips with 5 fps. 
The linear layer was trained for 
200 epochs on UCF101, 100 epochs on HMDB51, 35 epochs on K400 respectively.
For inference, we take the sliding windows along the time axis and 
apply spatial ten-crop strategy (4 corner crops and 1 center crop, with/without flipping),
and average the probability.
As shown in the paper Table~\ref{table:end_to_end},
after finetuning on the automatically-aligned HowTo100M subset,
the linear probe performances on all three datasets are clearly improved.
Note that, such procedure for alignment and end-to-end finetuning can be iteratively conducted 
to learn stronger video representations.
We leave these for future works.



\begin{figure*}[htb!]
	\centering
	\includegraphics[width=0.9\textwidth]{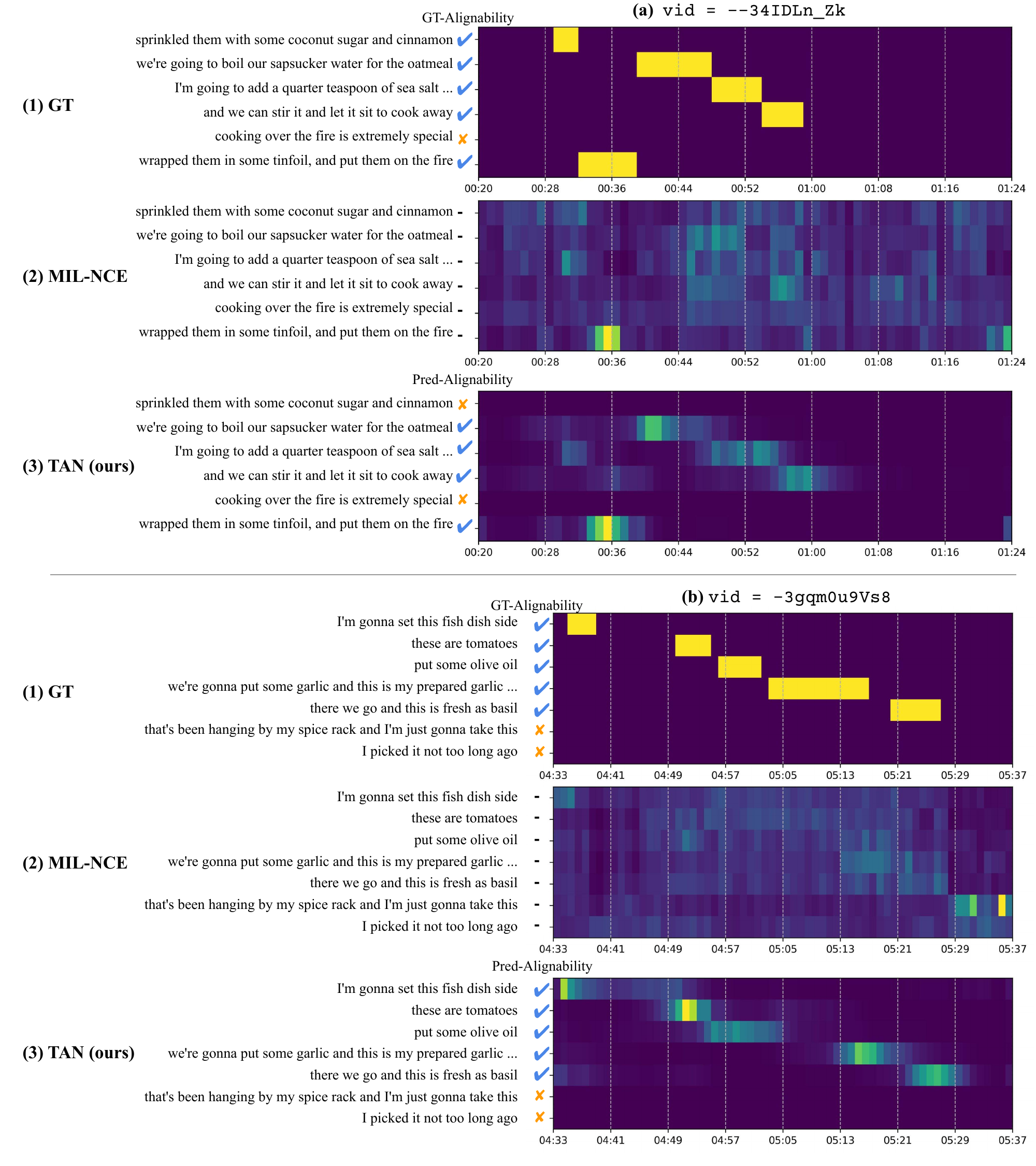}
	\vspace{-18pt}
	\caption{\small{Textual-visual heat-maps from MIL-NCE~\cite{Miech20} and our TAN, 
	computed on two examples from HTM-Align dataset.
	The raw sentence is shown on the y-axis.
	\textbf{(1)} The first row shows the ground-truth textual-visual alignment by manual annotation,
	which is a $K\times T$ \emph{binary} map where the lighter region means 
	the text is aligned to certain video segments. 
	The alignability of each text is also annotated.
	\textbf{(2)} The second row shows the $K\times T$ textual-visual similarity map 
	from the MIL-NCE backbone features. 
	Note that MIL-NCE does not classify the alignability of each sentence. 
	\textbf{(3)} The third row shows the $K\times T$ textual-visual similarity map ($\hat{\mathbb{A}}$)
	from our TAN. 
	Our model also predict the alignability of each sentence, which is shown on the figure.
\vspace{2pt}}}
\label{fig:heat}
\end{figure*}

\section{Qualitative Results}
In this section, we show two qualitative examples for temporal alignment,
and there are more detailed qualitative results in video format 
in our \emph{project webpage}.

Two examples from HTM-Align are shown in Figure~\ref{fig:heat}.
The first row shows the \emph{manually-aligned} timestamps of each sentence,
where some sentences are annotated as `not alignable'.
Note that in example (a) the ground-truth textual-visual alignment does not follow the monotonic temporal order,
this could happen in natural instructional videos
but DTW-type approaches cannot handle~\cite{Dvornik21}. 
This ordering issue is also discussed in the main paper Section~\ref{sec:intro}.
The second row shows the $K\times T$ heat-map from MIL-NCE~\cite{Miech20} backbone features.
The third row shows the alignment matrix $\hat{\mathbb{A}}$ from our Temporal Alignment Network.
For the clarity of visualization,
we normalise the similarity scores over the time axis with a softmax operation.

The qualitative results show that our method gives a much cleaner temporal alignment than MIL-NCE (row-3 vs row-2),
and our temporal alignment is close to human judgement (row-3 vs row-1).
The TAN also predicts an accurate alignability for each sentence that roughly matches human judgement.
In detail, the binary prediction is obtained by simply thresholding the probability with 0.5 threshold.
Our TAN is trained on the noisy HowTo100M as same as MIL-NCE,
but is able to learn a good temporal alignment by the proposed denoising mechanism,
showing the effectiveness of our method. 

\section{Limitations and Ethical Concerns}

Our work has the following limitations: 
(1) The current model can only process videos of about 1-minute long at a time, 
and is limited by the memory of existing GPUs hardware 
if we intend to align longer video sequence or train
the system end-to-end beyond this duration; 
(2) Our alignment method cannot handle repetitive text or repetitive visual content in a video
(for example in the video 
{\small \href{https://www.robots.ox.ac.uk/~vgg/research/tan/example.html}{\url{-0XlcSx1lNs.mp4}}} 
from our project webpage, 
the action `stir fry' occurs multiple times and is not aligned well), 
as one sentence can potentially be aligned to multiple video segments and vice versa. 
Note, this is a general problem in the tasks that require visual-language correspondence,
for example, in retrieval task~\cite{wray21}.
In practice, we find our proposed model does not suffer from this limitation though,
as repetitions are uncommon in natural instructional videos.

For ethical concerns, 
we are aware that we use a public instructional video dataset~\cite{Miech19} that uploaders shared on YouTube, 
which might have gender, age, geographical or cultural bias, 
also inevitably human faces appear in the dataset.

\end{document}